\documentclass{article}
\usepackage{amssymb}
\usepackage{graphicx}
\usepackage{amsfonts}
\usepackage{amsmath}
\usepackage{dsfont}
\usepackage[final]{corl_2025}
\usepackage{hyperref}

\title{Score the Steps, Not Just the Goal: VLM-Based Subgoal Evaluation for Robotic Manipulation}

\author{
  Ramy ElMallah\\
  Department of Mechanical and Industrial Engineering\\
  University of Toronto, Toronto, Canada\\
  \texttt{ramy@mie.utoronto.ca}
  \And
  Krish Chhajer\\
  Department of Electrical and Computer Engineering\\
  University of Toronto, Toronto, Canada\\
  \texttt{krish.chhajer@mail.utoronto.ca}
  \And
  Chi\mbox{-}Guhn Lee\\
  Department of Mechanical and Industrial Engineering\\
  University of Toronto, Toronto, Canada\\
  \texttt{cglee@mie.utoronto.ca}
}

\begin{document}
\maketitle


\begin{abstract}
Robot learning papers typically report a single binary success rate (SR), which obscures where a policy succeeds or fails along a multi-step manipulation task. We argue that subgoal-level reporting should become routine: for each trajectory, a vector of per-subgoal SRs that makes partial competence visible (e.g., grasp vs.\ pour). We propose a blueprint for StepEval, a cost-aware plug-in evaluation framework that utilizes vision–language models (VLMs) as automated judges of subgoal outcomes from recorded images or videos. Rather than proposing new benchmarks or APIs, our contribution is to outline design principles for a scalable, community-driven open-source project. In StepEval, the primary artifact for policy evaluation is the per-subgoal SR vector; however, other quantities (e.g., latency or cost estimates) are also considered for framework-optimization diagnostics to help the community tune evaluation efficiency and accuracy when ground-truth subgoal success labels are available. We discuss how such a framework can remain model-agnostic, support single- or multi-view inputs, and be lightweight enough to adopt across labs. The intended contribution is a shared direction: a minimal, extensible seed that invites open-source contributions, so that scoring the steps, not just the final goal, becomes a standard and reproducible practice.

\end{abstract}



\section{Introduction}
	
Robotic manipulation research largely evaluates policies with a single binary metric (success or failure). While simple to report, success rate (SR) provides a general, non-granular view of performance \citep{RoboLearningEmp, Kroemer2021-review}. A policy either achieves the task or not, with no insight into partial progress or which sub-task caused the failure. This is especially problematic for long-horizon tasks composed of multiple sequential subgoals (e.g., grasp, lift, place in a pick-and-place task). A low success rate might indicate failure, but it does not reveal where the policy struggled. As a motivating example, \citet{RoboLearningEmp} describe a pancake-flipping task consisting of several stages (picking up spatulas, flipping the pancake, then plating it). One policy in their study had only 17\% overall success, yet it completed the first two subgoals 100\% of the time; it consistently grasped the spatulas and flipped the pancake, failing only at the final plating step. If judged solely by overall SR, this policy would be deemed poor, obscuring the fact that it mastered over half the task. Granular evaluation is essential: without it, researchers cannot pinpoint failure modes (e.g., was it the grasp or the pour that failed?), making improvements guesswork.

The need for fine-grained, nuanced evaluation in robot learning has been increasingly recognized. Recent best-practice guidelines explicitly recommend reporting subgoal completion metrics and failure mode analyses alongside overall success \citep{RoboLearningEmp, roboeval, RoboCerebra}. However, adopting such practices in everyday research is challenging: manually labeling each sub-step success for dozens or hundreds of experiments is labor-intensive and subjective. Some works have automated subgoal checks in simulation (where the state is known) or relied on human labeling for physical trials. Vision-language models (VLMs) offer a promising middle ground, a way to automatically classify outcomes from visual data. Indeed, initial attempts have been made to use learned classifiers \citep{Inceoglu2023-eg, TaKSIE} or VLMs \citep{Guan2024-ki} to detect task failures.

Our goal is to reframe evaluation practice by advocating a subgoal-first view of manipulation metrics. We present StepEval as a blueprint for a cost-aware, plug-in framework that treats the per-subgoal success-rate (SR) vector as the primary artifact for policy evaluation. In this view, a VLM acts as a black-box automated judge that maps recorded imagery to subgoal outcomes. We articulate design principles for a community-built, scalable open-source project. All other quantities (e.g., latency, token/image cost, or confusion matrices based on ground-truth subgoal labels) are framed as framework-optimization diagnostics that help tune evaluation efficiency and accuracy when ground-truth labels are available, rather than as core evaluation metrics.

\medskip
We contribute by articulating the following pillars for subgoal-based evaluation:
\begin{itemize}
  \item \textbf{Subgoal-first evaluation perspective.} We formalize the evaluation target as a trajectory-level vector $\mathbf{y}\in\{0,1\}^n$ of per-subgoal SRs and argue that this granular object, rather than a single binary task SR, should become the standard reported outcome for manipulation studies.

  \item \textbf{Model-agnostic judging concept.} We propose using a VLM as a judge to infer subgoal outcomes from RGB image frames (single- or multi-view) without instrumenting policies or environments, keeping the framework post-hoc and lightweight. The design is intentionally policy- and model-agnostic, with the aim of making scoring the steps a routine, reproducible practice across labs.

  \item \textbf{Separation of evaluation vs.\ framework optimization metrics.} We distinguish the primary evaluation artifact (per-subgoal SR vector) from optional diagnostics (e.g., cost/latency; confusion matrices based on ground-truth subgoal labels; or optional prompt-optimization procedures). The latter are intended only to help the community tune efficiency and accuracy when labels exist, not to redefine the core evaluation target.
\end{itemize}

\section{Methodology}
In this section, we detail the formal problem setup and describe the design of the StepEval framework, including its architecture and usage workflow.
\subsection{Problem Setup}
We formalize the subgoal evaluation problem and the VLM-based solution here. Let task $T$ be composed of an ordered set of $n$ subgoals ${s_1, s_2, ..., s_n}$. Each subgoal $s_k$ is a definable unit of achievement (e.g., “robot grasps the object” for $s_1$). When a policy (controller) executes task $T$, it produces a trajectory $\tau$ which is a sequence of states or observations. In our vision-based setup, we consider $\tau$ as a sequence of images: $\tau = [I_1, I_2, ..., I_m]$ (these could be raw RGB frames from a single view or sets of images from multiple views at each timestep). For each subgoal $s_k$, there is an associated ground-truth outcome $y_k \in \{0,1\}$ indicating whether that subgoal was successfully completed (1) or not (0) in this trajectory. Hence each trajectory has an outcome vector $y = [y_1, ..., y_n] \in \{0,1\}^n$. The traditional success metric for the whole task is essentially $\mathds{1}[\min_k y_k = 1]$ (1 if all subgoals succeeded, 0 if any failed) – a very coarse projection of the full vector $y$. StepEval instead seeks to predict the full vector $y$ from the recorded trajectory, without manual labeling. Formally, we aim to learn or leverage a function $g: \tau \mapsto \hat{y}$ that maps a trajectory to a predicted success vector $\hat{y} \in {0,1}^n$ for all subgoals.

We propose to use a vision-language model (VLM) \(f_{\theta}\) to judge subgoal completion. In a simple prompting strategy we can issue one query per trajectory: given the subgoal set \(\mathcal{S}=\{s_1,\dots,s_n\}\) and the recorded trajectory \(\tau=[I_1,\dots,I_m]\), we form a prompt \(p(\mathcal{S},T)\) asking the VLM to indicate which subgoals were achieved, and we provide the entire visual context \(\mathbf{I}=\tau\). The model returns text that we parse into a binary vector
\[
\hat{\mathbf{y}}=(\hat{y}_1,\dots,\hat{y}_n)=f_{\theta}\!\left(p(\mathcal{S},T),\,\mathbf{I}\right)\in\{0,1\}^n,
\]
where \(\hat{y}_k=1\) denotes success of subgoal \(s_k\). The simple strategy does not attempt to isolate frames for individual subgoals; the VLM must infer outcomes from the global evidence in \(\tau\). Other potential prompting strategies are also possible, for example, prompting once per subgoal with \(p(s_k, T)\), using windowed or key-frame subsets \(I^{(k)}\subset\tau\).

We emphasize that $f_{\theta}$ is not trained on our task specifics; it might be a large foundation model like GPT-4o \cite{gpt4v-reasoning}, or an open model like GLM-4.5V \cite{glm}. These models have seen extensive vision-language data and can often answer questions about images. By tailoring the prompt, we engineer the model to act as a binary classifier for each subgoal. If needed, one could fine-tune a smaller model on labeled examples of successes/failures, but StepEval’s advantage is avoiding any task-specific training as we use the general knowledge in VLMs.

\subsection{Framework Architecture}

\begin{figure*}[t]
  \centering
  \includegraphics[width=\textwidth]{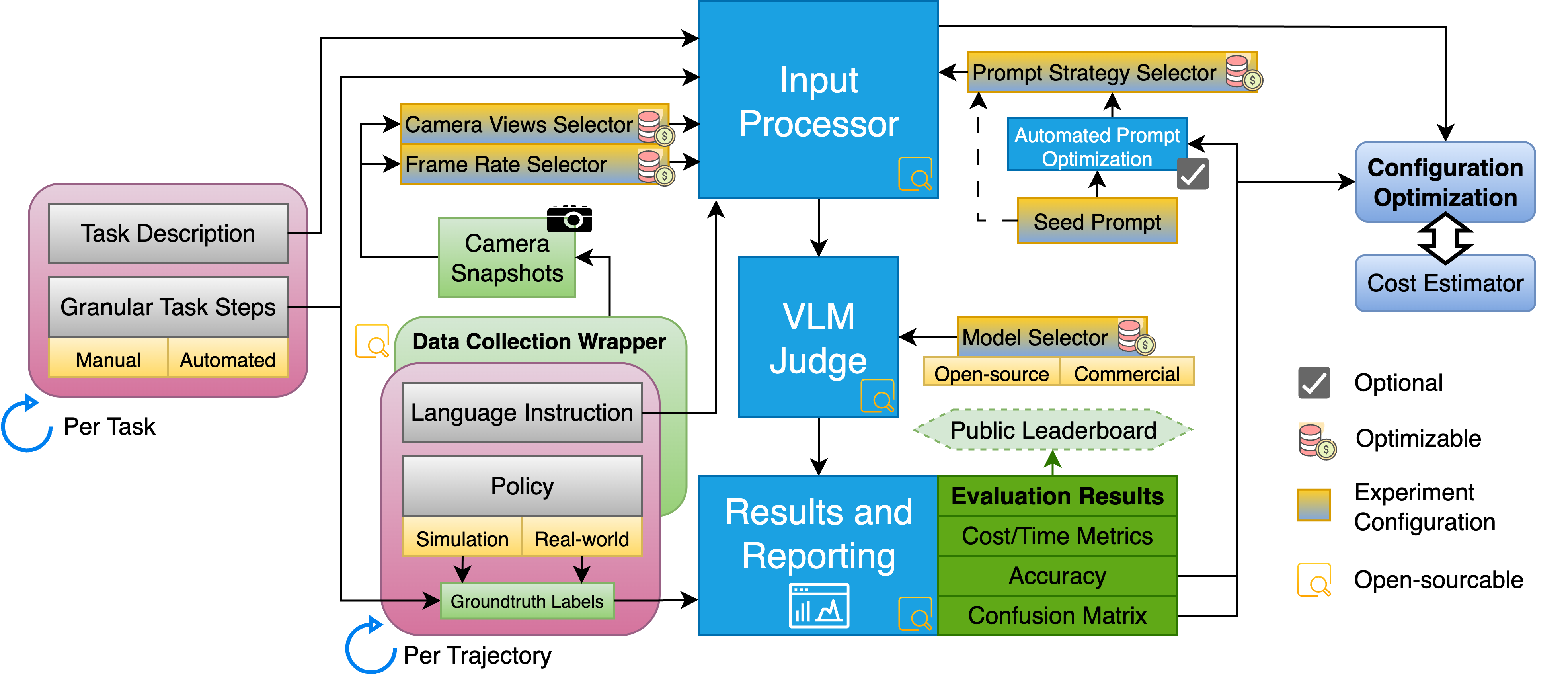}
  \caption{\textbf{Detailed Framework Development and Potential Open-source Contributions.}
  StepEval is modular. The \emph{Input Processor} compiles the input for the VLM judge based on the selected prompt, prompt strategy, camera view(s), and framerate.
  A \emph{Prompt Strategy} module supports templated prompts and optional automated prompt optimization for stronger VLM queries. 
  The \emph{VLM Judge} can be configured with either open-source models or commercial APIs. 
  \emph{Results \& Reporting} aggregates the evaluation results and optionally the framework-optimization metrics, including accuracy, confusion matrices, and cost/time, with an optional public-leaderboard export. 
  Around this core, a \emph{Data-Collection Wrapper} logs language instructions, runs policies in simulation or the real world, and saves camera snapshots.
  A \emph{Cost Estimator} and \emph{Configuration Optimization} loop help pick evaluation settings under time/budget constraints. 
  Modules marked as open-sourceable could be released by researchers to encourage community contributions and reproducible benchmarking.}
  \label{fig:stepeval-architecture}
\end{figure*}

The StepEval architecture, as shown in Figure \ref{fig:stepeval-architecture}, is designed as a seed that can grow into a scalable open-source project. The core objective is to produce, for each trajectory, a vector of per-subgoal outcomes usable as a primary evaluation artifact in manipulation studies. The design keeps evaluation post-hoc and policy agnostic: given recorded imagery and a subgoal rubric, a vision–language model acts as a black-box judge to infer subgoal outcomes. Surrounding this core, the framework exposes points where the community can plug in alternatives for prompt strategies, view and frame selection, and result reporting.

The input processor allows fusing multiple views to improve the visibility of subgoal evidence. Typical choices include front and wrist cameras, but the selector is not limited to these and depends on the available camera views in a specific experiment. Prompt strategies can be task-specific or generic and may use zero-shot or few-shot styles. Few-shot examples can improve judge accuracy in some settings at the cost of longer prompts; zero-shot prompts reduce overhead. These choices are configuration parameters rather than fixed implementations.

Results and reporting focus on the per-subgoal success vector. When ground-truth subgoal labels are available, additional diagnostics can be computed to characterize judge behavior and efficiency. The framework also allows optional tracking of computational or monetary cost and latency so that users can reason about trade-offs among models, views, and prompt choices. None of these diagnostics changes the primary evaluation goal.

\subsubsection{Judge accuracy diagnostics}
When ground-truth labels are available, agreement between predicted and true
subgoal outcomes can be summarized to assess a particular instantiation of the
judge and configuration.  Let $y_{j,k}\in\{0,1\}$ be the ground-truth outcome
for subgoal $k$ in trajectory $j$, and $\hat{y}_{j,k}$ the prediction.

Per-subgoal accuracy is
\[
A_k \;=\; \frac{1}{N}\sum_{j=1}^{N}
\mathds{1}\!\left[\hat{y}_{j,k}=y_{j,k}\right],
\]

and the task evaluation accuracy (i.e., how accurate the framework is in predicting the exact subgoals that have succeeded for a specific task) is
\[
A_{\text{task}}
\;=\;
\frac{1}{N}\sum_{j=1}^{N}
\mathds{1}\!
\Bigl[
\bigwedge_{k=1}^{n}
\bigl(\hat{y}_{j,k}=y_{j,k}\bigr)
\Bigr].
\]

Confusion matrices per subgoal can reveal systematic biases.  Because subgoals
are ordered, correlated errors may arise; for example, missed grasps can
propagate to later steps.  These diagnostics help select prompts, views, or
frame policies.

\subsubsection{Cost and latency diagnostics}

If a model provider charges $\alpha$ per 1{,}000 prompt tokens and $\beta$ per image, and the prompt evaluating a trajectory $k$ uses $T_k$ tokens and $M_k$ images, the expected cost per trajectory is
\[
C_k \;=\; \alpha \,\frac{T_k}{1000} \;+\; \beta\, M_k .
\]
These quantities allow users to compare configurations such as single- versus multi-view inputs or shorter versus longer prompts. They are intended to guide configuration under budget or time constraints. Latency can also be measured for each configuration. 

In summary, StepEval proposed architecture standardizes a post-hoc, model-agnostic pathway from recorded trajectories to per-subgoal outcome vectors while leaving implementation choices open for community iteration. Optional accuracy and efficiency diagnostics can be computed when labels exist to inform configuration choices, but the central object reported for policy evaluation remains the subgoal-level success vector.

\subsection{Usage Workflow}
StepEval is designed to be a four-step plug-in process for evaluating a manipulation policy at the subgoal level. The workflow is illustrated in Figure \ref{fig:stepeval-4steps}. A researcher can apply StepEval to any task as follows:

\begin{figure*}[t]
  \centering
  \includegraphics[width=\textwidth]{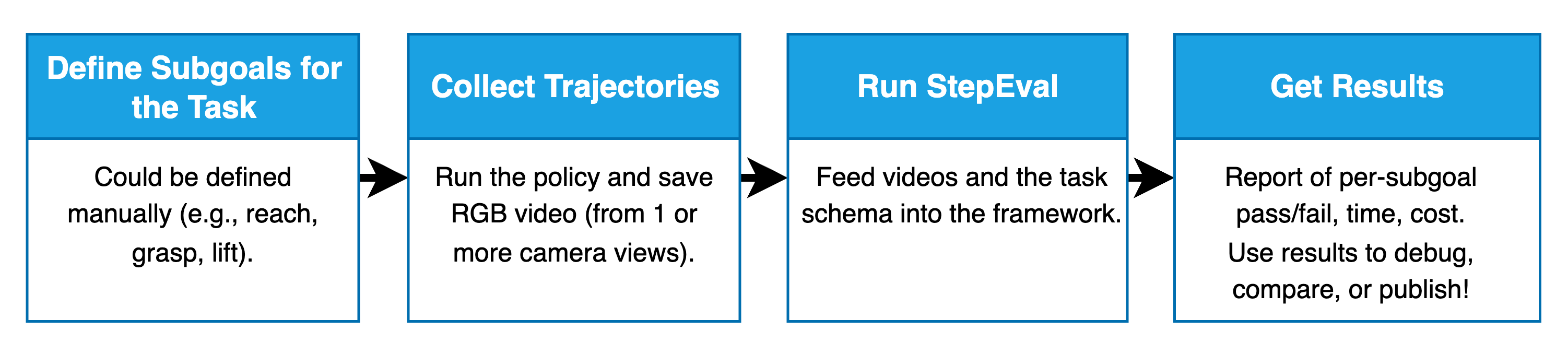}
  \caption{\textbf{Example Plug-and-Play Evaluation in 4 Steps: How Any Lab Can Benchmark a New Policy with Our Framework.}
  (1) Define the task’s subgoals. 
  (2) Collect policy rollouts.
  (3) Run the StepEval pipeline to classify each subgoal outcome using a vision–language model (VLM). 
  (4) Obtain a results report.}
  \label{fig:stepeval-4steps}
\end{figure*}

\textbf{1. Define Subgoals:} First, specify the task’s sequence of subgoals. This can be done manually (e.g., using domain knowledge to break down the task into meaningful steps: reach target, grasp object, lift object, etc.) or automatically (via the assistance of Multimodal LLM or using known task schemas). The output is a list of subgoal descriptions ${s_1, s_2, \dots, s_n}$ that the task comprises, in order. For example, a “Transfer Water” task might be segmented into [pick up cup], [align cup over bowl], [pour water], [place cup down]. Granular subgoals make evaluation more informative, but too many trivial subgoals can inflate complexity; therefore, a balance in granularity is chosen by the user or an automation module.

\textbf{2. Collect Trajectories:} Next, run the policy (controller or learned agent) to be evaluated and record its execution trajectories. In practice, this means capturing videos or image sequences from one or more cameras observing the scene. This step can be done in simulation or on real hardware. The only requirement is that the visual record is sufficient to judge each subgoal outcome. Optionally, if ground-truth labels for subgoal success are available (e.g., from simulation events or human labeling), these can be stored for later accuracy benchmarking of StepEval’s judgments. The output of this step is a set of trajectories ${\tau_j}$, where each $\tau$ is a sequence of RGB frames (or possibly multi-view frame sets) documenting one full attempt of the task.

\textbf{3. Run StepEval:} This is the core automated evaluation step. StepEval takes each trajectory and passes it through an Input Processor module, then into a VLM-based Judge, and finally produces evaluation results. The Input Processing involves configuring the input according to the requirements of the user. For example, this module can filter the image input based on the selected camera views (e.g., front view vs. wrist camera images) and frame rate. Next, the Prompt Strategy component prepares a textual prompt for the VLM. A few templated prompt options will be provided. If enabled, the automated prompt optimizer (e.g., using DSPy \citep{dspy}) can iteratively refine these prompts by slight paraphrasing or context augmentation, aiming to maximize agreement with known ground-truth labels on a validation set. Preliminary experiments suggest that automated prompt optimization can provide useful task-specific context when small labeled datasets are available. The chosen prompt $p$ and corresponding image(s) $I$ are then fed into the VLM (vision-language model) $f_{\theta}$. This model can be a large multimodal model, either open-source (run locally) or closed-source via an API. For each subgoal, the VLM outputs a judgment (success/failure decision). StepEval thus produces a predicted success vector $\hat{y} = [\hat{y}_1, \hat{y}_2, \dots, \hat{y}_n]$ for each trajectory, where $\hat{y}_k \in {0,1}$ indicates failure/success of subgoal $k$ as classified by the VLM.

\textbf{4. Get Results:} Finally, StepEval aggregates and reports the evaluation results. It generates a summary report including the per-subgoal success rate across all trajectories (how often each step was achieved), the overall task success rate (which could also be computed, but notably StepEval provides more detail than this single number), and a confusion matrix if ground-truth labels were provided (to analyze any misclassifications by the VLM judge). Crucially, the framework should allow computing cost metrics – e.g., the average time and monetary cost to evaluate one trajectory. Cost is primarily influenced by the chosen model (Self-hosted open-source models incur no API cost but might require GPU time; commercial models often have API pricing) and the number of prompts (which depends on the number of subgoals and prompt design). StepEval can thus output, for example, “Using Model X with single-view input, evaluating 100 trajectories cost \$5 and took 3 minutes, achieving 90\% subgoal classification accuracy.” This allows researchers to weigh the cost-benefit of different evaluation setups. Optionally, StepEval’s results can be exported in a standardized format or uploaded to a public leaderboard to compare subgoal success metrics across different policies or labs.

\section{Related Work}

\paragraph{LLMs/VLMs in Robotics.}
Early work grounded language models in robot affordances, using LLMs to choose feasible skills from high-level instructions \citep{SayCan}. Multimodal embodied LLMs scale this idea by ingesting images and robot states directly for long-horizon reasoning and planning \citep{PalmE}. Vision–language–action modeling pushes further by training a single model that maps visual–text inputs to action tokens, yielding strong zero-shot generalization \citep{rt2}. Off-the-shelf GPT-4V has also been used to extract stepwise plans from human demos and to bootstrap executable task programs \citep{Wake2023-wm}, while general prompting principles show how ChatGPT can be harnessed as a high-level planner \citep{chatgpt_robotics}. Scene-graph grounding improves scalability for building-size missions \citep{sayplan}. Open VLMs adapted for imitation (e.g., \citep{roboflamingo}) demonstrate that frozen multimodal backbones with lightweight policy heads can learn data-efficient manipulation. At the system level, embodied loops with GPT-4 enable long-horizon household tasks via retrieval-augmented planning and online replanning \citep{ellmer}. These works use LLMs/VLMs to act or plan. In contrast, StepEval uses VLMs as automated evaluators to score each subgoal post-hoc, is model-agnostic to the policy under test, and emphasizes cost-aware and plug-and-play deployment with prompt optimization rather than policy control.

\paragraph{Automated Failure Detection.}
Domain-tuned VLMs can recognize and explain manipulation failures from images (e.g., AHA \citep{aha}), trained on a large failure corpus generated via perturbations. RoboFAC pairs a vision-language-action model with a trajectory failure dataset to localize failure modes and suggest corrections that improve downstream success \citep{robofac}. SAFE predicts impending failure by reading the latent features of generalist VLA policies and calibrates decisions with conformal prediction \citep{safe}. VLMs can also supervise execution within Behavior Trees, detecting step-level failures and triggering reactive recovery \citep{Ahmad2025-qg}. Complementary “crash-testing’’ approaches proactively search for adversarial states that induce failures in a target policy \citep{robofail-robomd}. MLLM-IL \citep{mllm-il} proposes using MLLMs to automate the binary success detection of trajectories. Prior to the current VLM wave, multimodal fusion (vision + force) detected and classified failure types in real time, offering taxonomies of manipulation failure modes \citep{Inceoglu2023-eg}. These methods aim to detect, explain, or recover from failures during control. StepEval instead standardizes subgoal-level evaluation as an offline (or post-hoc) process that yields a binary success vector per trajectory. Our framework can plug in such detectors as judges, but our core contribution is a general, scalable, and inexpensive evaluation scaffold.

\paragraph{Robotic Policy Evaluation and Benchmarks.}
Meta-evaluation work urges the community to report statistics beyond binary SR (e.g., confidence intervals, robustness, safety) and to surface subgoal achievements and failure modes for transparency and reproducibility \citep{RoboLearningEmp}. Benchmarks such as ARNOLD \citep{arnold} probe language-grounded manipulation with continuous goal states and explicit generalization splits; The Colosseum \citep{COLOSSEUM} stresses systematic generalization across controlled perturbations with open leaderboards; SIMPLER \citep{simpler} provides high-fidelity sim testbeds that correlate with real-world evaluation for cost-effective assessment. Frameworks like RoboMimic \citep{robomimic} standardize imitation-learning evaluation across multi-task datasets and metrics. RoboEval \cite{roboeval} benchmarks bimanual manipulation with structured tasks and standardized diagnostics. Finally, \citet{Salfity2024-mw} propose subgoal segmentation and alignment metrics (temporal/semantic) to grade trajectories beyond pass/fail. These efforts provide guidelines, datasets, or testbeds; some reward partial credit or analyze generated code, and others quantify generalization. StepEval operationalizes the recommended practice of reporting subgoal outcomes by offering a VLM-based, black-box evaluator that turns raw videos into per-subgoal success vectors. Our focus is not on defining a new benchmark but on a portable evaluation tool that can be attached to any manipulation study (simulation or real), complementing existing benchmarks.


\section{Discussion and Conclusion}
\textbf{Advantages:} StepEval offers several clear benefits. First, it provides richer evaluation; instead of a one-bit success/fail outcome, researchers get a vector of outcomes for each sub-task, enabling pinpointing policy weaknesses. This is particularly useful for complex tasks or when comparing two policies: one can see exactly which subgoals differ in performance. Second, it is easy to adopt. After implementation, the proposed framework is plug-and-play with minimal integration: as long as you can record videos of your experiment, you can apply StepEval post-hoc. There is no need to instrument the robot or environment with additional sensors or to write custom evaluation code for each new task – a simple subgoal definition file suffices. Third, it is affordable and scalable. Using off-the-shelf models (especially open ones) means the cost per run is low, so one can evaluate hundreds of trials without worrying about annotation fatigue or budget. This could enable large-scale benchmarking of robot policies in a consistent way across labs (something that has been historically hard due to evaluation costs and protocol differences).

\textbf{Limitations:} Despite its promise, StepEval has some limitations. First, the accuracy of VLM judgments is inherently tied to what the camera can see. If a critical aspect of a subgoal is off-camera or ambiguous in the image (e.g., whether water actually flowed into the bowl might be hard to tell from one angle), the VLM may be unreliable. Careful camera placement or multi-view setups can mitigate this, but in practice, robots might not always have optimal viewpoints. We also assume the environment is similar to what the VLM has seen in training (common objects, etc.). If the scene is very novel (specialized lab equipment or poor lighting), the VLM’s reliability may drop, though big models are fairly robust in many settings. Another limitation is the need to define subgoals and success criteria manually. Camera resolution also introduces an accuracy–cost trade-off where lower resolutions can degrade recognition, whereas higher resolutions increase cost and latency. While this is flexible (the user can define what they care about), it does introduce some subjectivity and requires effort for each new task. In future work, we envision more automated ways to suggest subgoals and success criteria with the help of Multimodal LLMs. Additionally, current VLMs, even powerful ones, are not 100\% accurate. They can make errors, for example, misidentifying an object or producing a false positive result. However, model performance is rapidly improving.

\textbf{Future Vision:} We see StepEval as a starting point toward community-driven evaluation standards. One near-term vision is to create a public repository of subgoal evaluators for common benchmarks. For example, a user evaluating on CALVIN \citep{calvin} or ARNOLD \citep{arnold} would simply be able to download a configuration and run StepEval to obtain a complete report. Researchers could contribute improvements (new prompts, better models), which are then shared. Another idea is a leaderboard for policy performance, not just by overall score but by subgoal profile, encouraging reporting of, for instance, “[Method X] achieves 95\% overall success, struggling mainly at the final placing step (80\% success in that subgoal), whereas [Method Y] excels at placing but has occasional grasp failures.” Such a detailed comparison could spur targeted research to improve specific sub-skills. Over time, as this becomes common, reproducibility will improve, and future papers can compare where each method fails, not just how often it fails.

In conclusion, StepEval aims to make “scoring the steps” a natural part of robotics research, providing fine-grained insight at low cost. By leveraging the power of VLMs, we turn robotic manipulation policy evaluation, often the dull part of experiments, into something that is efficient, informative, and standardized. We invite the community to adopt StepEval, contribute to it, and collectively move toward more reproducible and rigorous evaluations in robot learning.


\clearpage


\bibliography{refs}  

\begin{thebibliography}{30}
\providecommand{\natexlab}[1]{#1}
\providecommand{\url}[1]{\texttt{#1}}
\expandafter\ifx\csname urlstyle\endcsname\relax
  \providecommand{\doi}[1]{doi: #1}\else
  \providecommand{\doi}{doi: \begingroup \urlstyle{rm}\Url}\fi

\bibitem[Kress-Gazit et~al.(2024)Kress-Gazit, Hashimoto, Kuppuswamy, Shah, Horgan, Richardson, Feng, and Burchfiel]{RoboLearningEmp}
H.~Kress-Gazit, K.~Hashimoto, N.~Kuppuswamy, P.~Shah, P.~Horgan, G.~Richardson, S.~Feng, and B.~Burchfiel.
\newblock Robot learning as an empirical science: Best practices for policy evaluation.
\newblock \emph{arXiv [cs.RO]}, Sept. 2024.

\bibitem[Kroemer et~al.(2021)Kroemer, Niekum, and Konidaris]{Kroemer2021-review}
O.~Kroemer, S.~Niekum, and G.~Konidaris.
\newblock A review of robot learning for manipulation: Challenges, representations, and algorithms.
\newblock \emph{J. Mach. Learn. Res.}, 22\penalty0 (30):\penalty0 1--82, 2021.

\bibitem[Wang et~al.(2025)Wang, Ung, Tannert, Duan, Li, Le, Oswal, Grotz, Pumacay, Deng, Krishna, Fox, and Srinivasa]{roboeval}
Y.~R. Wang, C.~Ung, G.~Tannert, J.~Duan, J.~Li, A.~Le, R.~Oswal, M.~Grotz, W.~Pumacay, Y.~Deng, R.~Krishna, D.~Fox, and S.~Srinivasa.
\newblock {RoboEval}: Where robotic manipulation meets structured and scalable evaluation.
\newblock \emph{arXiv [cs.RO]}, July 2025.

\bibitem[Han et~al.(2025)Han, Qiu, Liao, Huang, Gao, Yan, and Liu]{RoboCerebra}
S.~Han, B.~Qiu, Y.~Liao, S.~Huang, C.~Gao, S.~Yan, and S.~Liu.
\newblock {RoboCerebra}: A large-scale benchmark for long-horizon robotic manipulation evaluation.
\newblock \emph{arXiv [cs.RO]}, June 2025.

\bibitem[Inceoglu et~al.(2023)Inceoglu, Aksoy, and Sariel]{Inceoglu2023-eg}
A.~Inceoglu, E.~E. Aksoy, and S.~Sariel.
\newblock Multimodal detection and identification of robot manipulation failures.
\newblock \emph{arXiv [cs.RO]}, May 2023.

\bibitem[Kang and Kuo(2025)]{TaKSIE}
X.~Kang and Y.-L. Kuo.
\newblock Incorporating task progress knowledge for subgoal generation in robotic manipulation through image edits.
\newblock In \emph{2025 IEEE/CVF Winter Conference on Applications of Computer Vision (WACV)}, pages 7490--7499. IEEE, Feb. 2025.

\bibitem[Guan et~al.(2024)Guan, Zhou, Liu, Zha, Amor, and Kambhampati]{Guan2024-ki}
L.~Guan, Y.~Zhou, D.~Liu, Y.~Zha, H.~B. Amor, and S.~Kambhampati.
\newblock ``task success'' is not enough: Investigating the use of video-language models as behavior critics for catching undesirable agent behaviors.
\newblock \emph{arXiv [cs.AI]}, Feb. 2024.

\bibitem[Singh et~al.(2023)Singh, Cambronero, Gulwani, Le, and Verbruggen]{gpt4v-reasoning}
M.~Singh, J.~Cambronero, S.~Gulwani, V.~Le, and G.~Verbruggen.
\newblock Assessing {GPT4}-{V} on structured reasoning tasks.
\newblock \emph{arXiv [cs.CL]}, Dec. 2023.

\bibitem[{GLM-V Team} et~al.(2025){GLM-V Team}, Hong, Yu, Gu, Wang, Gan, Tang, Cheng, Qi, Ji, Pan, Duan, Wang, Wang, Cheng, He, Su, Yang, Pan, Zeng, Wang, Chen, Shi, Pang, Zhang, Yin, Yang, Chen, Xu, Zhu, Chen, Chen, Chen, Lin, Wang, Chen, Lei, Gong, Pan, Liu, Xu, Zhang, Zheng, Yang, Zhong, Huang, Zhao, Xue, Tu, Meng, Zhang, Luo, Hao, Tong, Li, Jia, Liu, Zhang, Lyu, Fan, Huang, Wang, Xue, Wang, Wang, An, Du, Shi, Huang, Niu, Wang, Yue, Li, Zhang, Wang, Wang, Zhang, Xue, Hou, Du, Wang, Zhang, Liu, Xu, Li, Huang, Dong, and Tang]{glm}
{GLM-V Team}, W.~Hong, W.~Yu, X.~Gu, G.~Wang, G.~Gan, H.~Tang, J.~Cheng, J.~Qi, J.~Ji, L.~Pan, S.~Duan, W.~Wang, Y.~Wang, Y.~Cheng, Z.~He, Z.~Su, Z.~Yang, Z.~Pan, A.~Zeng, B.~Wang, B.~Chen, B.~Shi, C.~Pang, C.~Zhang, D.~Yin, F.~Yang, G.~Chen, J.~Xu, J.~Zhu, J.~Chen, J.~Chen, J.~Chen, J.~Lin, J.~Wang, J.~Chen, L.~Lei, L.~Gong, L.~Pan, M.~Liu, M.~Xu, M.~Zhang, Q.~Zheng, S.~Yang, S.~Zhong, S.~Huang, S.~Zhao, S.~Xue, S.~Tu, S.~Meng, T.~Zhang, T.~Luo, T.~Hao, T.~Tong, W.~Li, W.~Jia, X.~Liu, X.~Zhang, X.~Lyu, X.~Fan, X.~Huang, Y.~Wang, Y.~Xue, Y.~Wang, Y.~Wang, Y.~An, Y.~Du, Y.~Shi, Y.~Huang, Y.~Niu, Y.~Wang, Y.~Yue, Y.~Li, Y.~Zhang, Y.~Wang, Y.~Wang, Y.~Zhang, Z.~Xue, Z.~Hou, Z.~Du, Z.~Wang, P.~Zhang, D.~Liu, B.~Xu, J.~Li, M.~Huang, Y.~Dong, and J.~Tang.
\newblock {GLM}-4.{5V} and {GLM}-4.{1V}-thinking: Towards versatile multimodal reasoning with scalable reinforcement learning.
\newblock \emph{arXiv [cs.CV]}, Aug. 2025.

\bibitem[Khattab et~al.(2023)Khattab, Singhvi, Maheshwari, Zhang, Santhanam, Vardhamanan, Haq, Sharma, Joshi, Moazam, Miller, Zaharia, and Potts]{dspy}
O.~Khattab, A.~Singhvi, P.~Maheshwari, Z.~Zhang, K.~Santhanam, S.~Vardhamanan, S.~Haq, A.~Sharma, T.~T. Joshi, H.~Moazam, H.~Miller, M.~Zaharia, and C.~Potts.
\newblock {DSPy}: Compiling declarative language model calls into self-improving pipelines.
\newblock \emph{arXiv [cs.CL]}, Oct. 2023.

\bibitem[Ahn et~al.(2022)Ahn, Brohan, Brown, Chebotar, Cortes, David, Finn, Fu, Gopalakrishnan, Hausman, Herzog, Ho, Hsu, Ibarz, Ichter, Irpan, Jang, Ruano, Jeffrey, Jesmonth, Joshi, Julian, Kalashnikov, Kuang, Lee, Levine, Lu, Luu, Parada, Pastor, Quiambao, Rao, Rettinghouse, Reyes, Sermanet, Sievers, Tan, Toshev, Vanhoucke, Xia, Xiao, Xu, Xu, Yan, and Zeng]{SayCan}
M.~Ahn, A.~Brohan, N.~Brown, Y.~Chebotar, O.~Cortes, B.~David, C.~Finn, C.~Fu, K.~Gopalakrishnan, K.~Hausman, A.~Herzog, D.~Ho, J.~Hsu, J.~Ibarz, B.~Ichter, A.~Irpan, E.~Jang, R.~J. Ruano, K.~Jeffrey, S.~Jesmonth, N.~J. Joshi, R.~Julian, D.~Kalashnikov, Y.~Kuang, K.-H. Lee, S.~Levine, Y.~Lu, L.~Luu, C.~Parada, P.~Pastor, J.~Quiambao, K.~Rao, J.~Rettinghouse, D.~Reyes, P.~Sermanet, N.~Sievers, C.~Tan, A.~Toshev, V.~Vanhoucke, F.~Xia, T.~Xiao, P.~Xu, S.~Xu, M.~Yan, and A.~Zeng.
\newblock Do as {I} can, not as {I} say: Grounding language in robotic affordances.
\newblock \emph{arXiv [cs.RO]}, Apr. 2022.

\bibitem[Driess et~al.(2023)Driess, Xia, Sajjadi, Lynch, Chowdhery, Ichter, Wahid, Tompson, Vuong, Yu, Huang, Chebotar, Sermanet, Duckworth, Levine, Vanhoucke, Hausman, Toussaint, Greff, Zeng, Mordatch, and Florence]{PalmE}
D.~Driess, F.~Xia, M.~S.~M. Sajjadi, C.~Lynch, A.~Chowdhery, B.~Ichter, A.~Wahid, J.~Tompson, Q.~Vuong, T.~Yu, W.~Huang, Y.~Chebotar, P.~Sermanet, D.~Duckworth, S.~Levine, V.~Vanhoucke, K.~Hausman, M.~Toussaint, K.~Greff, A.~Zeng, I.~Mordatch, and P.~Florence.
\newblock {PaLM}-{E}: An embodied multimodal language model.
\newblock \emph{arXiv [cs.LG]}, Mar. 2023.

\bibitem[Brohan et~al.(2023)Brohan, Brown, Carbajal, Chebotar, Chen, Choromanski, Ding, Driess, Dubey, Finn, Florence, Fu, Arenas, Gopalakrishnan, Han, Hausman, Herzog, Hsu, Ichter, Irpan, Joshi, Julian, Kalashnikov, Kuang, Leal, Lee, Lee, Levine, Lu, Michalewski, Mordatch, Pertsch, Rao, Reymann, Ryoo, Salazar, Sanketi, Sermanet, Singh, Singh, Soricut, Tran, Vanhoucke, Vuong, Wahid, Welker, Wohlhart, Wu, Xia, Xiao, Xu, Xu, Yu, and Zitkovich]{rt2}
A.~Brohan, N.~Brown, J.~Carbajal, Y.~Chebotar, X.~Chen, K.~Choromanski, T.~Ding, D.~Driess, A.~Dubey, C.~Finn, P.~Florence, C.~Fu, M.~G. Arenas, K.~Gopalakrishnan, K.~Han, K.~Hausman, A.~Herzog, J.~Hsu, B.~Ichter, A.~Irpan, N.~Joshi, R.~Julian, D.~Kalashnikov, Y.~Kuang, I.~Leal, L.~Lee, T.-W.~E. Lee, S.~Levine, Y.~Lu, H.~Michalewski, I.~Mordatch, K.~Pertsch, K.~Rao, K.~Reymann, M.~Ryoo, G.~Salazar, P.~Sanketi, P.~Sermanet, J.~Singh, A.~Singh, R.~Soricut, H.~Tran, V.~Vanhoucke, Q.~Vuong, A.~Wahid, S.~Welker, P.~Wohlhart, J.~Wu, F.~Xia, T.~Xiao, P.~Xu, S.~Xu, T.~Yu, and B.~Zitkovich.
\newblock {RT}-2: Vision-language-action models transfer web knowledge to robotic control.
\newblock \emph{arXiv [cs.RO]}, July 2023.

\bibitem[Wake et~al.(2023)Wake, Kanehira, Sasabuchi, Takamatsu, and Ikeuchi]{Wake2023-wm}
N.~Wake, A.~Kanehira, K.~Sasabuchi, J.~Takamatsu, and K.~Ikeuchi.
\newblock {GPT}-{4V}(ision) for robotics: Multimodal task planning from human demonstration.
\newblock \emph{arXiv [cs.RO]}, Nov. 2023.

\bibitem[Vemprala et~al.(2023)Vemprala, Bonatti, Bucker, and Kapoor]{chatgpt_robotics}
S.~Vemprala, R.~Bonatti, A.~Bucker, and A.~Kapoor.
\newblock {ChatGPT} for robotics: Design principles and model abilities.
\newblock \emph{arXiv [cs.AI]}, Feb. 2023.

\bibitem[Rana et~al.(2023)Rana, Haviland, Garg, Abou-Chakra, Reid, and Suenderhauf]{sayplan}
K.~Rana, J.~Haviland, S.~Garg, J.~Abou-Chakra, I.~Reid, and N.~Suenderhauf.
\newblock {SayPlan}: Grounding large language models using {3D} scene graphs for scalable robot task planning.
\newblock \emph{arXiv [cs.RO]}, July 2023.

\bibitem[Li et~al.(2023)Li, Liu, Zhang, Yu, Xu, Wu, Cheang, Jing, Zhang, Liu, Li, and Kong]{roboflamingo}
X.~Li, M.~Liu, H.~Zhang, C.~Yu, J.~Xu, H.~Wu, C.~Cheang, Y.~Jing, W.~Zhang, H.~Liu, H.~Li, and T.~Kong.
\newblock Vision-language foundation models as effective robot imitators.
\newblock \emph{arXiv [cs.RO]}, Nov. 2023.

\bibitem[Mon-Williams et~al.(2025)Mon-Williams, Li, Long, Du, and Lucas]{ellmer}
R.~Mon-Williams, G.~Li, R.~Long, W.~Du, and C.~G. Lucas.
\newblock Embodied large language models enable robots to complete complex tasks in unpredictable environments.
\newblock \emph{Nat. Mach. Intell.}, 7\penalty0 (4):\penalty0 592--601, Mar. 2025.

\bibitem[Duan et~al.(2024)Duan, Pumacay, Kumar, Wang, Tian, Yuan, Krishna, Fox, Mandlekar, and Guo]{aha}
J.~Duan, W.~Pumacay, N.~Kumar, Y.~R. Wang, S.~Tian, W.~Yuan, R.~Krishna, D.~Fox, A.~Mandlekar, and Y.~Guo.
\newblock {AHA}: A vision-language-model for detecting and reasoning over failures in robotic manipulation.
\newblock \emph{arXiv [cs.RO]}, Sept. 2024.

\bibitem[Lu et~al.(2025)Lu, Ye, Ye, Tao, Yang, and Zhao]{robofac}
W.~Lu, M.~Ye, Z.~Ye, R.~Tao, S.~Yang, and B.~Zhao.
\newblock {RoboFAC}: A comprehensive framework for robotic failure analysis and correction.
\newblock \emph{arXiv [cs.RO]}, 2025.

\bibitem[Gu et~al.(2025)Gu, Ju, Sun, Gilitschenski, Nishimura, Itkina, and Shkurti]{safe}
Q.~Gu, Y.~Ju, S.~Sun, I.~Gilitschenski, H.~Nishimura, M.~Itkina, and F.~Shkurti.
\newblock {SAFE}: Multitask failure detection for vision-language-action models.
\newblock \emph{arXiv [cs.RO]}, June 2025.

\bibitem[Ahmad et~al.(2025)Ahmad, Ismail, Styrud, Stenmark, and Krueger]{Ahmad2025-qg}
F.~Ahmad, H.~Ismail, J.~Styrud, M.~Stenmark, and V.~Krueger.
\newblock A unified framework for real-time failure handling in robotics using vision-language models, reactive planner and behavior trees.
\newblock \emph{arXiv [cs.RO]}, 2025.

\bibitem[Sagar et~al.(2024)Sagar, Duan, Vasudevan, Zhou, Amor, Fox, and Senanayake]{robofail-robomd}
S.~Sagar, J.~Duan, S.~Vasudevan, Y.~Zhou, H.~B. Amor, D.~Fox, and R.~Senanayake.
\newblock From mystery to mastery: Failure diagnosis for improving manipulation policies.
\newblock \emph{arXiv [cs.RO]}, Dec. 2024.

\bibitem[ElMallah et~al.(2024)ElMallah, Zamani, and Lee]{mllm-il}
R.~ElMallah, N.~Zamani, and C.-G. Lee.
\newblock Human 0, {MLLM} 1: Unlocking new layers of automation in language-conditioned robotics with multimodal {LLMs}.
\newblock In \emph{2024 21st International Conference on Mechatronics - Mechatronika (ME)}, pages 1--8. IEEE, Dec. 2024.

\bibitem[Gong et~al.(2023)Gong, Huang, Zhao, Geng, Gao, Wu, Ai, Zhou, Terzopoulos, Zhu, Jia, and Huang]{arnold}
R.~Gong, J.~Huang, Y.~Zhao, H.~Geng, X.~Gao, Q.~Wu, W.~Ai, Z.~Zhou, D.~Terzopoulos, S.-C. Zhu, B.~Jia, and S.~Huang.
\newblock {ARNOLD}: A benchmark for language-grounded task learning with continuous states in realistic {3D} scenes.
\newblock \emph{arXiv [cs.AI]}, Apr. 2023.

\bibitem[Pumacay et~al.(2024)Pumacay, Singh, Duan, Krishna, Thomason, and Fox]{COLOSSEUM}
W.~Pumacay, I.~Singh, J.~Duan, R.~Krishna, J.~Thomason, and D.~Fox.
\newblock {THE} {COLOSSEUM}: A benchmark for evaluating generalization for robotic manipulation.
\newblock \emph{arXiv [cs.RO]}, Feb. 2024.

\bibitem[Li et~al.(2024)Li, Hsu, Gu, Pertsch, Mees, Walke, Fu, Lunawat, Sieh, Kirmani, Levine, Wu, Finn, Su, Vuong, and Xiao]{simpler}
X.~Li, K.~Hsu, J.~Gu, K.~Pertsch, O.~Mees, H.~R. Walke, C.~Fu, I.~Lunawat, I.~Sieh, S.~Kirmani, S.~Levine, J.~Wu, C.~Finn, H.~Su, Q.~Vuong, and T.~Xiao.
\newblock Evaluating real-world robot manipulation policies in simulation.
\newblock \emph{arXiv [cs.RO]}, May 2024.

\bibitem[Mandlekar et~al.(2021)Mandlekar, Xu, Wong, Nasiriany, Wang, Kulkarni, Fei-Fei, Savarese, Zhu, and Martín-Martín]{robomimic}
A.~Mandlekar, D.~Xu, J.~Wong, S.~Nasiriany, C.~Wang, R.~Kulkarni, L.~Fei-Fei, S.~Savarese, Y.~Zhu, and R.~Martín-Martín.
\newblock What matters in learning from offline human demonstrations for robot manipulation.
\newblock \emph{arXiv [cs.RO]}, Aug. 2021.

\bibitem[Salfity et~al.(2024)Salfity, Wanna, Choi, and Pryor]{Salfity2024-mw}
J.~Salfity, S.~Wanna, M.~Choi, and M.~Pryor.
\newblock Temporal and semantic evaluation metrics for foundation models in post-hoc analysis of robotic sub-tasks.
\newblock \emph{arXiv [cs.RO]}, Mar. 2024.

\bibitem[Mees et~al.(2021)Mees, Hermann, Rosete-Beas, and Burgard]{calvin}
O.~Mees, L.~Hermann, E.~Rosete-Beas, and W.~Burgard.
\newblock {CALVIN}: A benchmark for language-conditioned policy learning for long-horizon robot manipulation tasks.
\newblock \emph{arXiv [cs.RO]}, Dec. 2021.

\end{thebibliography}

\end{document}